\documentclass[11pt,onecolumn,journal]{IEEEtran}
\usepackage{amsmath,amsfonts}
\usepackage{algorithmic}
\usepackage{algorithm}
\usepackage{array}
\usepackage[caption=false,font=normalsize,labelfont=sf,textfont=sf]{subfig}
\usepackage{textcomp}
\usepackage{stfloats}
\usepackage{url}
\usepackage{verbatim}
\usepackage{graphicx}
\usepackage{cite}
\usepackage{hyperref}
\hypersetup{
    colorlinks = true,
    citecolor = darkblue,
    linkcolor = black
}
\usepackage{booktabs}

\usepackage{siunitx}

\usepackage{microtype}     %
\usepackage{calc} %
\usepackage{xcolor}         %
\usepackage{bm}         %
\usepackage{graphicx}  		%

\usepackage{amsmath} %
\usepackage{cite} %
\usepackage{diagbox}
\usepackage{multirow}
\usepackage{tabularx}
\usepackage{hhline}

\usepackage{arydshln}
\usepackage{mathtools}
\usepackage{amsthm}	
\usepackage{amssymb}
\usepackage{pifont}
\usepackage{threeparttable}
\usepackage{outlines}

\usepackage{arydshln}
\usepackage[export]{adjustbox}

\usepackage{mathtools} 
\usepackage{microtype} 
\usepackage{multirow}
\usepackage{nicefrac}       
\usepackage{pifont}   

\usepackage{tabularx}
\usepackage{threeparttable} 
\usepackage{caption}
\usepackage{tabularray}
\UseTblrLibrary{booktabs}
\usepackage{wrapfig}
\usepackage{lipsum}
\usepackage{setspace}
\usepackage{tikz}

\usepackage{makecell} %

\definecolor{darkblue}{rgb}{0,0 ,0.542}
\definecolor{lightgreen}{rgb}{.9,1,.9}
\definecolor{lightred}{rgb}{1,.415,.415}
\definecolor{lightblue}{rgb}{.415,.415,1}

\usepackage[letterpaper, left=1.in, right=1.in, top=1in, bottom=1in]{geometry}

\renewcommand{\vec}[1]{\bm{#1}} 

\newcommand{\X}{{\vec{X}}}  %
\newcommand{\Y}{{\vec{Y}}}  %

\newcommand{\N}{{\vec{N}}}  %

\def\x{\vec{x}}  %
\def\y{\vec{y}}  %
\def\s{\vec{s}}  %

\newcommand{\fX}{f_{\X}}
\newcommand{\fXsigma}{f_{\X_\sigma}}

\newcommand{\normpdf}{{\mathcal{N}}}  

\def\argmin{\mathop{\mathsf{arg\,min}}} %

\begin{document}

\title{Random Walks with Tweedie:\\[0.3em]
A Unified View of Score-Based Diffusion Models\\[0.3em]
}
\author{%
\normalsize \textbf{Chicago~Y.~Park}$^1$ \quad \textbf{Michael~T.~McCann}$^2$ \quad \textbf{Cristina~Garcia-Cardona}$^2$ \quad \\[0.7ex] \textbf{Brendt~Wohlberg}$^2$ \quad \textbf{Ulugbek~S.~Kamilov}$^1$\\[0.7em]
\small \textnormal{$^1$Washington University in St. Louis} \quad \quad \textnormal{$^2$Los Alamos National Laboratory} \\[0.5em]
\footnotesize \texttt{\{chicago,  kamilov\}@wustl.edu}\\[0.7ex]
\footnotesize \texttt{\{mccann,  cgarciac, brendt\}@lanl.gov}
}

\maketitle

\begin{abstract}
\label{sec:abstract}
\noindent
 We present a concise derivation for several influential score-based diffusion models
that relies on only a few textbook results.
Diffusion models have recently emerged as powerful tools for generating realistic, synthetic signals --- %
particularly natural images --- %
and often play a role in state-of-the-art algorithms for inverse problems in image processing.
While these algorithms are often surprisingly simple,
the theory behind them is not,
and multiple complex theoretical justifications exist in the literature.
Here,
we provide a simple and largely self-contained theoretical justification
for score-based diffusion models that is targeted towards the signal processing community.
This approach leads to generic algorithmic templates for training and generating samples with diffusion models.
We show that several influential diffusion models correspond to particular choices within these templates and demonstrate that alternative, more straightforward algorithmic choices can provide comparable results.
This approach has the added benefit of enabling conditional sampling without any likelihood approximation.
\end{abstract}

\section{Introduction}
\label{sec:Introduction}
This paper focuses on 
deep score-based diffusion models,
which are a type of generative model that uses a deep neural network to learn a score function --- %
the gradient of the logarithm of a probability density --- %
from training data.
This score function is then used in an iterative algorithm
based on the mathematical theory of diffusion
to generate novel samples.
These models have produced state-of-the-art results in generating natural images~\cite{dhariwal2021beat}, videos~\cite{ho2022videodiffusion}, and 3D objects~\cite{zhang20233dsceneSOTA} (see \cite{po2024sotareviewsurvey} for a recent survey).
These algorithms have also been adapted to generate samples conditioned on noisy measurements,
enabling their use for solving inverse problems (see \cite{chung2024reviewofdiffusion} for a recent survey).

There are several popular score-based diffusion model algorithms that involve training a neural network for denoising and then generating samples by starting from random noise and progressively denoising it through iterative steps.
Despite this superficial resemblance, they 
are rooted in seemingly distinct mathematical theories.
The noise conditional score network (NCSN)\cite{song2019generative} justifies training a diffusion model using the result that the denoising objective is equivalent to explicit score matching, as shown in\cite{vincent_connection_2011},
and justifies its iterative sampling via Langevin dynamics,
without providing a reference that explains why it results in the desired distribution.
The denoising diffusion probabilistic model (DDPM)~\cite{ho_NEURIPS2020_ddpm} explicitly defines the generative model as a finite discrete-time Markov chain with learned Gaussian transitions,
leading to iterative sampling;
it shows that training a denoiser is equivalent to optimizing a weighted variational bound on likelihood for this model.
The score-based generative modeling through stochastic differential equations framework (SGM-SDE)~\cite{song2021sde}
provides a generalization of NCSN and DDPM.
It defines a forward stochastic differential equation (SDE), interprets sampling as solving a corresponding reverse SDE
and justifies the denoising training objective in the same way as NCSN.

The main aim of this work is to present a simple derivation of diffusion models
that enables viewing the distinctions between different diffusion model algorithms as algorithmic choices
rather than as essential features of the underlying mathematical theory.
The result is a template for training and a template for sampling,
which is both mathematically justified and general enough to encompass NCSN~\cite{song2019generative}, 
DDPM~\cite{ho_NEURIPS2020_ddpm},
and SGM-SDE~\cite{song2021sde},
see Table~\ref{table:template_training} and Table~\ref{table:template_sampling}.
It is also flexible in that it allows the noise schedule during training to differ from the one used during sampling
and
is compatible with pretrained denoisers.
Finally,
the framework enables conditional sampling with a simple application of Bayes' rule, without any likelihood approximation or conditional training.
While none of the individual components of the theory described here are new,
we believe that presenting them together provides a uniquely succinct and approachable exposition.
For each component, we have aimed to provide both a standard textbook reference
and
instances of similar ideas in the diffusion model literature.
The code for reproducing unconditional and conditional sampling results in the paper is publicly available.\footnote{\href{https://github.com/wustl-cig/randomwalk_diffusion}{\textcolor{magenta}{https://github.com/wustl-cig/randomwalk\_diffusion}}}

The paper is organized as follows.
Section~\ref{sec:Introduction} introduces score-based diffusion models and outlines the motivation for a unified theoretical framework.
Section~\ref{sec:theory_diffusion} presents the core score-based diffusion theory in two steps: (1) linking MMSE denoising to the score via Tweedie’s formula, and (2) using the score to define the sequence of random walks for sampling.
Section~\ref{sec:proposed_templates} shows how this leads to general templates that can include several influential score-based diffusion models.
Section~\ref{sec:theory_diffusion_conditional} extends the framework to conditional sampling without likelihood approximations.
Section~\ref{sec:numerical_validation} demonstrates unconditional and conditional generation using our approach.

\begin{table}[t]
    \setstretch{1.5} %
    \centering
    \caption{Proposed training template and parameter choices.}
    \renewcommand{\arraystretch}{1.5}
    \begin{tabular}{@{}llll@{}}
    \toprule
    \noalign{\vskip -1.5ex}
    \textbf{Training template} &
    \multicolumn{3}{@{}l}{ $\mathcal{L}(\theta) =
    \mathbb{E}_{\sigma, \X, \N} \left[ w(\sigma) \| \X - \mathsf{D}_{\theta} (\X + \sigma\N, \sigma) \|^2_2 \right], \quad \X \sim \fX$} \\ 
    \cmidrule{1-4}\\ \noalign{\vskip -8.0ex}
      & \multicolumn{1}{c}{\multirow{1}{*}{$\sigma$}} & \multicolumn{1}{c}{\multirow{1}{*}{$w(\sigma)$}} & \multicolumn{1}{c}{\multirow{1}{*}{$\mathsf{D}_{\theta}(\x, \sigma)$}}  
     \\ [-1.2ex]
    \cmidrule{2-4} 
    \textbf{NCSN}~\cite{song2019generative}, \textbf{VE SDE}~\cite{song2021sde}  &
    \multicolumn{1}{c}{\multirow{1}{*}{$\sim \mathcal{U}([\sigma_{\mathrm{min}}, \sigma_{\mathrm{max}})]$}} &
    \multicolumn{1}{c}{\multirow{1}{*}{$\sigma^{-4}$}} & \multicolumn{1}{c}{\multirow{1}{*}{$\x + \sigma^2\s_\theta(\x, \sigma)$}} \\

    \textbf{DDPM}~\cite{ho_NEURIPS2020_ddpm}, \textbf{VP SDE}~\cite{song2021sde} &
    \multicolumn{1}{c}{\multirow{1}{*}{$\sqrt{(1 - \bar{\alpha}_{t})/{\bar{\alpha}_{t}}}, \, t \sim \mathcal{U}(\{1,2, \dots, T\})$}} & 
    
    \multicolumn{1}{c}{\multirow{1}{*}{$\sigma^{-2}$}} &
    \multicolumn{1}{c}{\multirow{1}{*}{$ \x - \sigma \bm{\epsilon}_\theta \left(\sqrt{\bar{\alpha}_{t(\sigma)}} \x, t(\sigma)\right)$}}\\
    \bottomrule
    \end{tabular}
    \label{table:template_training}
\end{table}

\subsection{Related Work}

There is a growing literature trying to simplify, explain, or unify diffusion model theory.
A variational inference perspective has been introduced~\cite{luo2022googlediffusiontutorial, chan2024tutorial}, framing diffusion models as a special case of Markovian hierarchical variational autoencoders.
This view shows that diffusion models can be derived using the Evidence Lower Bound (ELBO), as in standard variational autoencoders, unifying multi-step diffusion processes with hierarchical latent variable models under probabilistic inference.
Another recent perspective~\cite{kingma2023understandingdiffusionELBO} shows that diffusion training objectives can be interpreted as ELBOs, under the condition that the weighting function is monotonic and the noise is added as Gaussian augmentation.
A structured explanation of DDPM~\cite{ho_NEURIPS2020_ddpm} has been proposed by breaking down the model into six intuitive design steps, each clarifying how standard generative modeling is reframed as a sequence of supervised denoising tasks~\cite{turner2024ddpmsixsteps}.
Simplification of score training has been proposed by learning the score of the velocity conditioned on data --- %
under Hamiltonian dynamics --- %
instead of directly learning the score from the data~\cite{dockhorn2022cldsgm}.

Several other works share our goal of unifying some aspects of the theory of generative diffusion models.
The mathematical equivalence between time-inhomogeneous discrete-time diffusion models and time-homogeneous continuous-time Ornstein-Uhlenbeck processes evaluated at non-uniformly distributed observation times was demonstrated in \cite{santos2023ornstein_ddpm}.
A design-oriented formulation~\cite{karras2022edm} has been proposed to explain how differences between score-based sampling algorithms arise from the choice of scaling applied to the mean and noise variance in the marginal data distribution, along with the input and output transformations used in the score model.
The stochastic interpolant framework~\cite{albergo2023stochasticinterpolant} introduces a unified formulation for flow- and diffusion-based generative models by representing sample trajectories between arbitrary distributions as a combination of drift and tunable noise components.
A drift-decomposition framework~\cite{pandey2023completerecipediffusion} has been proposed to parameterize forward SDEs using a score-driven term and a skew-symmetric rotational component, capturing SGM-SDE~\cite{song2021sde} as a special case.

\begin{table}[t]
    \setstretch{1.5} %
    \centering
    \caption{Proposed sampling template and parameter choices.}
    \renewcommand{\arraystretch}{1.5}
    \begin{tabular}{@{}llll@{}}
        \toprule
        \noalign{\vskip -1.25ex}
          \textbf{Sampling template} &
            \multicolumn{3}{c}{$
    \x_{k+1} = \x_k + \tau_k \underbrace{\frac{\mathsf{D}_{\theta}(\x_k, \sigma_k) -\x_k }{\sigma_k^2}}_{\textstyle
    \nabla \log f_{\vec{X}_{\sigma_{k}}}(\x_k)}
    + \sqrt{2\tau_k \mathcal{T}_k} \vec{n}$ 
    }\\
    \cmidrule{1-4}\\ \noalign{\vskip -8.0ex}

    & \multicolumn{1}{c}{\multirow{1}{*}{$\sigma_k$}}  & \multicolumn{1}{c}{\multirow{1}{*}{$\tau_k$}} & \multicolumn{1}{c}{\multirow{1}{*}{$\mathcal{T}_k$}}
     \\ [-1.2ex]
    \cmidrule{2-4} 
    \textbf{NCSN}~\cite{song2019generative} &  
    \multicolumn{1}{c}{\multirow{1}{*}{$\sigma_k$}} & \multicolumn{1}{c}{\multirow{1}{*}{$\varepsilon\sigma_k^2 /(2\sigma_{\mathrm{max}}^2)$}} & \multicolumn{1}{c}{\multirow{1}{*}{$1$}}
    \\
    \textbf{VE SDE}~\cite{song2021sde} 
    & \multicolumn{1}{c}{\multirow{1}{*}{$\sigma_k$}} & \multicolumn{1}{c}{\multirow{1}{*}{$\sigma_k^2 - \sigma_{k+1}^2$}} & \multicolumn{1}{c}{\multirow{1}{*}{$1/2$}}
    \\

    \textbf{DDPM}~\cite{ho_NEURIPS2020_ddpm}, \textbf{VP SDE}~\cite{song2021sde}
    &
    \multicolumn{1}{c}{\multirow{1}{*}{$\sqrt{(1-\bar{\alpha}_{T-k})/\bar{\alpha}_{T-k}}$}} &
     \multicolumn{1}{c}{\multirow{1}{*}{$(1-\alpha_{T-k})/\bar{\alpha}_{T-k}$}} &
     \multicolumn{1}{c}{\multirow{1}{*}{${\sigma'}_{T-k}^2\alpha_{T-k}/(2-2\alpha_{T-k})$}}
    \\
    \bottomrule
    \end{tabular}
    \label{table:template_sampling}
\end{table}

\section{Theory of Deep Score-Based Diffusion Sampling}
\label{sec:theory_diffusion}

We now present our theoretical justification for
deep score-based generative modeling.
While these results are not new
(in fact, we have intentionally relied on textbook results),
when brought together
they offer a simple and mathematically rigorous justification
for many diffusion model algorithms.

\subsection{Tweedie's Formula}
\label{subsec:tweedie}
The score function --- %
the gradient of the logarithm of a probability density function (PDF) --- %
guides the denoising process in diffusion sampling, leading noisy data toward areas of higher probability density for image sampling.
Tweedie's formula~\cite{efron2011tweedie} establishes a relationship between the score function and an optimal denoiser, offering an elegant approximation of the score using only the MMSE reconstruction and noisy images.

We denote a noisy version of a random variable, $\X \sim \fX$,
by
\begin{equation} \label{eq:X_sigma}
    \X_\sigma = \X + \sigma\N, 
    \quad
    \N \sim \normpdf(\vec{0}, \vec{I})\;,
\end{equation}
where $\sigma$ is the specified noise level, and $\vec{N}$ is a standard multivariate Gaussian.

An MMSE reconstruction function is defined as one that recovers the image that, on average, is the closest to the ground truth given the observed measurements
\begin{equation}
\label{equ:x_hat_mmse}
    \hat{\x}_{\text{MMSE}}(\x) = \argmin_{\hat{\x}} \mathbb{E} \left[ \|\hat{\x} - \X\|_2^2 \mid \vec{X}_{\sigma} = \x \right]\;.
\end{equation}
Tweedie's formula establishes the relationship between the score function and the MMSE denoiser as
\[
\nabla \log \fXsigma (\x) = 
    \frac{\hat{\x}_{\text{MMSE}}(\x) - \x}{\sigma^2}
\]
for $\x \in \mathbb{R}^{d}$.

Note that $\fXsigma$ is the noisy image distribution, which is distinct from the desired image distribution $\fX$. Since the sum of independent random variables leads to a distribution that is the convolution of their individual distributions, $\fXsigma$ is given by 
\[
\fXsigma(\x) = \int_{\mathbb{R}^{d}} \fX(\x') \phi_{\sigma}(\x-\x') d\x'\;,
\]
where $\phi_{\sigma}(\cdot)$ is the probability density function of $\sigma\N$. %
Thus, we can view $\fXsigma$ as a smoothed version of $\fX$,
and we can see from \eqref{eq:X_sigma} that (roughly speaking) $\nabla \log \fXsigma$ approaches $\nabla \log \fX$ as $\sigma$ approaches $0$.

While Tweedie’s formula is implicitly involved in denoising score-matching diffusion models as seen in \cite{song2019generative, song2021sde, ho_NEURIPS2020_ddpm}, this connection was made explicit in \cite{mccann2023score, daras2024ambienttweedie}.

\subsection{Pretrained Denoisers as MMSE Estimators}
\label{subsec:proof_cnn_mmse}

Tweedie's formula relates score functions with MMSE denoisers;
we now show that neural networks trained on a denoising objective
approximate MMSE denoisers.
Consider the denoising inverse problem, where $\X_{\sigma} = \X + \sigma \N$ is a random variable representing $\X$ perturbed by Gaussian noise with variance $\sigma^2$, and denote the corresponding MMSE denoiser by 
$\hat{\x}_{\text{MMSE}}: \mathbb{R}^{d} \to \mathbb{R}^{d}$.

Given a set of independent and identically distributed (IID) training samples, $\X^{(1)}, \X^{(2)}, \dots, \X^{(N)} \sim \fX$,
we form noisy versions according to $\X_{\sigma}^{(n)} = \X^{(n)} + \sigma \N$.
Then the mean squared error (MSE) loss for a reconstruction algorithm, $\hat{\x}: \mathbb{R}^{d} \to \mathbb{R}^{d}$, is
\begin{equation} \label{eq:loss_alg_mmse}
\begin{aligned}
    L(\hat{\x}) &= \frac{1}{N} \sum_{n=1}^{N} w(\sigma)\|\hat{\x}(\X_{\sigma}^{(n)}) - \X^{(n)}\|_2^2 \\
            &\approx \mathbb{E} \left[w(\sigma)\|\hat{\x}(\X_{\sigma}) - \X\|_2^2\right]\;,
\end{aligned}
\end{equation}
where $w(\sigma) > 0$ is user-defined weighting factor.
The best possible reconstruction algorithm according to the average MSE metric is exactly the MMSE estimator \eqref{equ:x_hat_mmse} because
\begin{subequations} \label{eq:MMSE_proof}
\begin{align}
    \hat{\x}_{\text{MMSE}}(\x) &= \argmin_{\hat{\x}} \mathbb{E} \left[ \|\hat{\x} - \X\|_2^2 \mid \X_{\sigma} = \x \right]\\
     &= \argmin_{\hat{\x}} \int  \|\hat{\x} - \x'\|_2^2 f_{\X|\X_{\sigma}}(\x'|\x)d\x' \\
     &= \argmin_{\hat{\x}} \int  \|\hat{\x} - \x'\|_2^2 \frac{f_{\X,\X_{\sigma}}(\x',\x)}{f_{\X_{\sigma}}(\x)}d\x' \\
     &= \argmin_{\hat{\x}} \int  \|\hat{\x} - \x'\|_2^2 f_{\X,\X_{\sigma}}(\x',\x)d\x'\;,
\end{align}
\end{subequations}
which implies that
\[
    \int  \|\hat{\x}_{\text{MMSE}}(\x)  - \x'\|_2^2 f_{\X,\X_{\sigma}}(\x',\x)d\x' \leq \int  \|\hat{\x}(\x)  - \x'\|_2^2 f_{\X,\X_{\sigma}}(\x',\x)d\x'
\]
for any reconstruction algorithm $\hat{\x}$ and noisy image $\x$.
Integrating both sides over $\x$
and incorporating the positive weighting factor $w(\sigma)$  gives 
\[
    \mathbb{E}\left[ w(\sigma)\|\hat{\x}_{\text{MMSE}}(\X_{\sigma})-\X\|^2_2 \right] \leq \mathbb{E}\left[ w(\sigma)\|\hat{\x}(\X_{\sigma})-\X\|^2_2)\right]
\]
for all $\hat{\x}$.
This implies that, in the limit of an infinitely large test set, 
an MMSE image reconstruction algorithm, by definition, achieves the lowest average MSE among all possible algorithms.

Thus, a well-trained deep denoiser \( \mathsf{D}_{\theta} \) --- %
optimized to minimize the loss function in \eqref{eq:loss_alg_mmse} --- %
can serve as an effective approximation of the MMSE estimator.
The score can then be derived from this denoiser using Tweedie's formula
\[
\nabla \log \fXsigma (\x) = 
    \frac{\mathsf{D}_{\theta}(\x) - \x}{\sigma^2}\;.
\]

Similar ideas have appeared in \cite{vincent2011connection}, which
proposed estimating the score through denoising based on the principle that the gradient of the log density at a noisy sample $\X_{\sigma}$ naturally points toward the noise-free sample $\X$.
The connection between MSE loss and MMSE estimation is derived in the textbook~\cite{chan_introduction_2021}.

\subsection{Random Walk in a Potential and the Score Function}
\label{subsec:fokker_planck}

Building on textbook results from stochastic differential equation theory, we show that simulating a particular random walk allows us to sample a PDF while only having access to the corresponding score function.

The motion (in $d$ dimensions) of a particle influenced by Brownian motion
and a potential $V$ can be described by the equation
\begin{equation} \label{eq:langevin_diffusion}
    d\X_{t} = -\nabla V(\X_{t}) dt + \sqrt{2\mathcal{T}} d\vec{W}_{t}\;,
\end{equation}
where 
$\X_{t}$ is a random variable that represents the position of the particle at time $t$,
$V: \mathbb{R}^d \rightarrow \mathbb{R}$ represents a potential, $\vec{W}_{t}$ denotes standard Brownian motion on $\mathbb{R}^d$, 
the temperature, $\mathcal{T}\in \mathbb{R},  \mathcal{T}> 0$, controls the extent of the random fluctuations,
and $\X_{0}$ represents the initial distribution~\cite[Section 3.5, pg. 73]{pavliotis2014langevintextbook}.

If $V$ grows fast enough that the particle cannot escape to infinity
(see \cite[Definition 4.2]{pavliotis2014langevintextbook} for the technical condition),
then
for any distribution $f_{\X_{0}}$ of initial positions,
the distribution
\[
    f_{\mathcal{T}}(\x) = \frac{1}{Z} e^{-\frac{V(\x)}{\mathcal{T}}}\;,
\]
where
\[
    Z = \int_{\mathbb{R}^d} e^{-\frac{V(\x)}{\mathcal{T}}} d\x
\]
is the unique invariant distribution corresponding to \eqref{eq:langevin_diffusion}
\cite[Proposition 4.2]{pavliotis2014langevintextbook}.
This means that
\[
    \lim_{t \to \infty} f_{\X_{t}} = f_{\mathcal{T}}(\x) \;.
\]

As a first step towards our goal of drawing samples from a target distribution $\fX$,
we choose the potential
 \[
 V(\x) = -\log \fX(\x)\;.
 \]

Equation \eqref{eq:langevin_diffusion} can be discretized and solved using the Euler-Maruyama method~\cite[Section 5.2, pg. 146]{pavliotis2014langevintextbook} as
\begin{equation}
 \label{eq:Euler-Maruyama}
    \x_{k+1} = \x_{k} + 
     \tau_{k}\nabla \log \fX(\x_{k})
    + \sqrt{2\tau_{k}\mathcal{T}} \vec{n},\quad k \in  \{0, 1, \ldots, K-1\}\;,
\end{equation}
where $\nabla \log \fX(\x_{k})$ is the score function, $\tau_{k}$ denotes the step size at iteration $k$, $\vec{n} \sim \mathcal{N}(\vec{0}, \vec{I})$, and $\mathcal{T}$ is the temperature
(which we will later generalize to a sequence $\{\mathcal{T}_k\}$).
Starting from any initial point $\x \in \mathbb{R}^d$, this algorithm converges to a solution of \eqref{eq:langevin_diffusion} in the limit of
$\tau_k$ small enough and $K$ large enough~\cite{leobacher2018eulerconvergence}.

\subsection{Decreasing the Noise Level During Sampling}
\label{subsec:seq_random_walks}

We have thus far shown (closely following~\cite{mccann2023score})  how a trained score can be used to generate samples from a noisy distribution, $\X_\sigma$.
In practice, diffusion model algorithms decrease $\sigma$ during sampling
so as to generate samples that closely approximate the (non-noisy) target distribution
$\X_{\sigma = 0}=\X$.
In the discussion section, we consider possible alternatives to this approach.

In the current framework,
changing the noise level $\sigma$ during sampling
corresponds to solving 
a sequence of $K$ random walks with different potential functions,
\begin{equation} \label{eq:sequence_random_walks}
    d\x_{t} =
    \begin{cases}
    \nabla \log f_{\X_{\sigma_0}}(\x_{t}) dt + \sqrt{2 \mathcal{T}_0} d\vec{W}_{t} & t\in [0, t(\sigma_0))\\
    \nabla \log f_{\X_{\sigma_1}}(\x_{t}) dt + \sqrt{2 \mathcal{T}_1} d\vec{W}_{t} & t \in [t(\sigma_0), t(\sigma_1)) \\
    \mathrel{\makebox[\widthof{$V_1(\x_{t}) dt + \sqrt{2 \mathcal{T}} d\vec{W}_{t}$}]{\vdots}} & 
    \mathrel{\makebox[\widthof{$t \in [t_0, t_1)$}]{\vdots}}\\ 
    \nabla \log f_{\X_{\sigma_{K-1}}}(\x_{t}) dt + \sqrt{2 \mathcal{T}_{K-1}} d\vec{W}_{t} & t \in [t(\sigma_{K-2}), t(\sigma_{K-1})) \\
    \end{cases}
\end{equation}
where \( t(\sigma_k) \) denotes the timestep associated with noise level \( \sigma_k \).
The noise levels $\sigma_k$ decrease from a large value at $k=0$ to a small value at $\sigma_{K-1}$
so that $f_{\X_{\sigma_{K-1}}} \approx \fX$ (Taking $\sigma_{K-1}=0$ in this framework results in a division by zero in Tweedie's formula).
Or, alternatively, solving a random walk with a continuously varying potential,
\begin{equation}
    d\x_{t} = \nabla \log f_{\X_{\sigma_t}}(\x_{t}) dt + \sqrt{2 \mathcal{T}_t} d\vec{w}_{t}\;.
\end{equation}
A version of this idea appeared in \cite{song2019generative} under the name annealed Langevin dynamics.

Because the solver \eqref{eq:Euler-Maruyama} converges regardless of where it is initialized,
the sequence of random walks \eqref{eq:sequence_random_walks} can theoretically be solved
by applying \eqref{eq:Euler-Maruyama}  directly to the last equation
and taking a sufficient number of small-enough steps.
In practice,
we instead obtain a numerical solution to \eqref{eq:sequence_random_walks} by running
a sequence of numerical solvers,
each initialized with the result of the previous one.
A heuristic justification for this approach is as follows.
If $\sigma$ decreases smoothly enough
(i.e., if the number of potential functions, $K$, is large relative to the difference between $\sigma_0$ and $\sigma_{K-1}$), 
our solvers will converge to a solution
faster than the underlying SDE will change.
Taking this idea to its limit,
we can take a single step of the solver for each $\sigma_t$.
Doing so leads to our proposed sampling template,
which we describe in the next section.
(For in-depth convergence analysis of diffusion models, see~\cite{debortoli2021schrodingerbridge, lee2022diffusionconvergence2, debortoli2022diffusionconvergence3}.)

\section{Proposed Diffusion Model Templates}\label{sec:proposed_templates}
In summary,
a diffusion model algorithm can be derived by following a training and sampling template.
The training template is
\begin{equation} \label{eq:template_loss}
    \mathcal{L}(\theta) =
    \mathbb{E}_{\sigma, \X, \X_\sigma} \left[ w(\sigma) \| \X - \mathsf{D}_{\theta} (\X_\sigma, \sigma) \|^2_2\right]\;,
\end{equation}
where $w : \mathbb{R}_{\ge0} \to \mathbb{R}_{\ge0}$ is a (user-specified) weight function,
$\X \sim \fX$ is a draw from the target distribution,
$\sigma \sim f_\sigma$ is a user-specified noise scale distribution,
and
$\X_\sigma = \X + \sigma \N$ is a noisy version of $\X$.
Similar templates have been derived elsewhere, e.g.,
in \cite{kingma2023understandingdiffusionELBO}, where it is derived as a weighted variational lower bound.

The sampling template is
\begin{equation} \label{eq:template_score} %
    \x_{k+1} = \x_k + \tau_k\nabla \log f_{\vec{X}_{\sigma_{k}}}(\x_k) + \sqrt{2\tau_k \mathcal{T}_k} \vec{n},\quad k \in  \{0, 1, \ldots, K-1\}\;,
\end{equation}
where the sequences $\{\sigma_k\}$, \{$\tau_k$\}, and \{$\mathcal{T}_k$\} are algorithmic parameters that
represent noise levels, step sizes, and temperatures, respectively.
From a theoretical perspective, the choice of the initial value $\x_0$ is arbitrary,
but practically, it should be similar to a sample from $\X_{\sigma_0}$
because the network $\mathsf{D}_{\theta}(\cdot, \sigma_0)$ was trained on these samples.
The noise level $\sigma_k$ should decrease towards zero as the iteration $k$ approaches $K-1$.
The step size $\tau_k$ controls the balance between numerical stability
(increasing with smaller $\tau$)
and the number of steps required for good quality samples
(decreasing with larger $\tau$).
The temperature sequence $\mathcal{T}_k$ should end with $\mathcal{T}_{K-1}=1$ for sampling to be mathematically correct;
setting the temperature below one drives sampling towards high-density regions,
setting it larger than one promotes faster mixing.
We explore this effect in Section~\ref{subsec:temperaturecontrol} and Figure~\ref{fig:different_temperature}.
Note that the $\x_k$'s in \eqref{eq:template_score} are random variables;
while we have used uppercase bold symbols to denote random variables so far, 
we have used lowercase for consistency with existing formulations.

When training is performed with the training template,
we can substitute the trained network for the score function,
\begin{equation} \label{eq:template_network} %
    \x_{k+1} = \x_k + \frac{\tau_k}{\sigma_k^2}\left(\mathsf{D}_{\theta}(\x_k, \sigma_k) -\x_k \right)   + \sqrt{2\tau_k \mathcal{T}_k} \vec{n},\quad k \in  \{0, 1, \ldots, K-1\}\;,
\end{equation}
or equivalently
\[
    \x_{k+1} = \left(1-\frac{\tau_k}{\sigma_k^2} \right)\x_k + \left(\frac{\tau_k}{\sigma_k^2}\right) \mathsf{D}_{\theta}(\x_k, \sigma_k)   + \sqrt{2\tau_k \mathcal{T}_k} \vec{n},\quad k \in  \{0, 1, \ldots, K-1\}\;.
\]
The first form resembles a randomized gradient descent and provides the intuition that $\tau_k$ should be chosen to be small relative to $\sigma_k^2$ to keep the effective step size small.
The second form emphasizes that the iterates mix the current image $\x_k$ with its denoised version $\mathsf{D}_{\theta}(\x_k)$.

In the following sections,
we describe how the sampling template may be used with pretrained denoisers
and interpret several existing diffusion model algorithms
in terms of our proposed templates.

\subsection{Using Pretrained Denoisers}
\label{sec:any_denoisers}
Our sampling template admits the use
of pretrained denoisers, however, care must be taken to properly scale their inputs and outputs to conform to the template.
One way to achieve this is to interpret the training procedure in terms of the training template \eqref{eq:template_loss}.
Doing so results in an expression for the denoiser in the template, $\mathsf{D}_{\theta}$,
in terms of the pretrained denoiser,
which can then be used in the denoiser form of the sampling template~\eqref{eq:template_network};
this is the approach we take in Section~\ref{sec:interp}.
In other cases, it is easier to work in terms of the score function; we pursue this approach here.

In deep score-based diffusion sampling,
variance-exploding (VE) and variance-preserving (VP) score functions (and denoising networks) are widely used.
These score functions are defined by two different ways of adding noise.
In the VE setting, which NCSN and VE-SDE adopt, the noise variance increases at each step, with more noise being added as the process progresses according to
\[
    \X^{\text{VE}}_{t} = \X + \sigma_t\vec{N}\;.
\]
In the VP setting, which DDPM and VP-SDE adopt,
the total variance of the data is kept constant throughout the noise addition,
\begin{equation} \label{equ:vp_noise_perturbation}
    \X^{\text{VP}}_{t} = \sqrt{\bar{\alpha}_t}
    \cdot
    \left(
    \X + \sqrt{\frac{1-\bar{\alpha}_t}{\bar{\alpha}_t}}\cdot\vec{N}
    \right)\;,
\end{equation}
where \(\bar{\alpha}_t = \prod_{s=1}^t \alpha_s\), with $t =1, 2, \dots, T$. Here, \(\alpha_t\) is chosen to ensure that \(\X^{\text{VP}}_0\) follows the desired probability distribution and \(\X^{\text{VP}}_T\) follows a known distribution, such as a standard Gaussian.

Due to the constant scale term applied to the noise-free image $\X$ in \eqref{equ:vp_noise_perturbation},
the VE score cannot be directly substituted for the VP score, and vice versa.
However, the VE noise perturbation can be expressed in terms of the VP one: $\X^{\text{VE}}_{t} = (1/\sqrt{\bar{\alpha}_t})\X^{\text{VP}}_{t}$ when $\sigma_t$ is defined as $\sigma_t = \sqrt{(1-\bar{\alpha}_t)/\bar{\alpha}_t}$.
For a probability density function scaled by different random variables, we can derive the gradient (i.e., score) relationship as
\[
    f_{\X^{\text{VE}}_{t}}(\x) = \sqrt{\bar{\alpha}_t}f_{\X^{\text{VP}}_{t}}(\sqrt{\bar{\alpha}_t}\x)\;,
\]
where \(\sigma_t = \sqrt{(1 - \bar{\alpha}_t)/\bar{\alpha}_t}\), we proceed by taking the logarithm on both sides and differentiate with respect to $\x$:
\begin{equation} \label{equ:vp_score_and_ve_score}
    \nabla \log f_{\X^{\text{VE}}_{t}}(\x) = \sqrt{\bar{\alpha}_t} \nabla \log f_{\X^{\text{VP}}_{t}}(\sqrt{\bar{\alpha}_t}\x)\;.
\end{equation}

Comparing  $\X^{\text{VE}}_{t}$ with $\X_{\sigma}$ in \eqref{eq:X_sigma},
we can derive:
\[
    \begin{aligned}
    \nabla \log f_{\X_{\sigma_t}}(\x) &= \nabla \log f_{\X^{\text{VE}}_{t}}(\x) \\
            &= \sqrt{\bar{\alpha}_t} \nabla \log f_{\X^{\text{VP}}_{t}}(\sqrt{\bar{\alpha}_t}\x)\;,
    \end{aligned}
\]
which shows how to make networks trained via the VE and VP approach compatible with the sampling template \eqref{eq:template_score}.
This score relationship also allows us to relate VE and VP scores via the MMSE reconstruction of the noise-free image \( \mathsf{D}_{\theta} \):
\[
\begin{aligned}
    \mathsf{D}_{\theta}(\x_t) &= \x_t + \sigma_t^2\nabla \log f_{\X^{\text{VE}}_{t}}(\x_t) \\
            &= \x_t + \frac{1-\bar{\alpha}_t}{\sqrt{\bar{\alpha}_t}}\nabla \log f_{\X^{\text{VP}}_{t}}(\sqrt{\bar{\alpha}_t}\x_t)\;.
\end{aligned}
\]

\subsection{Interpretation of Existing Methods} \label{sec:interp}

\textbf{The VP training objective in our template.} DDPM~\cite{ho_NEURIPS2020_ddpm} and VP-SDE~\cite{song2021sde} use the same VP training objective.
To represent this objective in terms of the training template \eqref{eq:template_loss},
we set the noise variance distribution to
\begin{equation} \label{eq:VP_training_sigma}
    \sigma = \sqrt{\frac{
1 - \bar{\alpha}_{t}
    }{
\bar{\alpha}_{t}
    }} \quad t \sim \text{Uniform}({1, \dots, T})\;,
\end{equation}
with $\bar{\alpha}_{t}$ defined as in \cite{ho_NEURIPS2020_ddpm}.
Let $w(\sigma) = \sigma^{-2}$.
Let the denoiser $\mathsf{D}_{\theta}$ take the form 
\begin{equation}
    \mathsf{D}_{\theta} (\x, \sigma) = 
    \x 
    -     \sigma \bm{\epsilon}_\theta(\sqrt{\bar{\alpha}_{t(\sigma)}} \x, t(\sigma))\;,
\end{equation}
where $t(\sigma)$ denotes the conditional time input $t$ that corresponds to a given $\sigma$ via \eqref{eq:VP_training_sigma}
and $\vec{\epsilon}_{\theta}$ represents a neural network trained to predict the noise component of 
$\x$ given its scaled input, as introduced in \cite{ho_NEURIPS2020_ddpm}.

Substituting these choices into the template loss, we have
\begin{subequations}
\begin{align}
    \eqref{eq:template_loss} &= 
    \mathbb{E}_{\sigma, \X, \X_\sigma} \left[ \sigma^{-2} \| \X - (\X_\sigma - \sigma \bm{\epsilon}_\theta(\sqrt{\bar{\alpha}_{t(\sigma)}} \X_\sigma, t(\sigma) )) \|^2_2 \right] \\
    &=  \mathbb{E}_{\sigma, \X, \N}  \left[ \sigma^{-2} \| \X - (\X + \sigma \N - \sigma \bm{\epsilon}_\theta(\sqrt{\bar{\alpha}_{t(\sigma)}} \X + \sqrt{\bar{\alpha}_{t(\sigma)}} \sigma \N, t(\sigma) )) \|^2_2 \right]  \\
    &=  \mathbb{E}_{\sigma, \X, \N}  \left[ \|  \N -  \bm{\epsilon}_\theta(\sqrt{\bar{\alpha}_{t(\sigma)}} \X + \sqrt{\bar{\alpha}_{t(\sigma)}} \sigma \N, t(\sigma) ) \|^2_2 \right]     \\
    &=  \mathbb{E}_{t, \X, \N}  \left[ \|  \N -  \bm{\epsilon}_\theta(\sqrt{\bar{\alpha}_{t}} \X + \sqrt{\bar{\alpha}_{t}} \sigma \N, t) \|^2_2 \right]     \\    
    &=  \mathbb{E}_{t, \X, \N}  \left[ \|  \N -  \bm{\epsilon}_\theta(\sqrt{\bar{\alpha}_t} \X + \sqrt{1-\bar{\alpha}_t}\N, t ) \|^2_2 \right]\;,
\end{align}
\end{subequations}
which is the VP training loss \cite[Algorithm 1 and (14)]{ho_NEURIPS2020_ddpm}.

\textbf{The VE training objective in our template.} NCSN~\cite{song2019generative} and VE-SDE~\cite{song2021sde} both use the same VE training objective.
To represent this objective in terms of the training template \eqref{eq:template_loss}:
Let $\{\sigma_{\ell}\}_{\ell=0}^{L-1}$ be a decreasing geometric sequence with $\sigma_0$ as the largest term and $\sigma_{L-1}$ as the smallest, satisfying $\frac{\sigma_0}{\sigma_1} = \dots = \frac{\sigma_{L-2}}{\sigma_{L-1}} > 1$.
Let the denoiser take the form
\[
    \mathsf{D}_{\theta} (\x, \sigma) = 
    \x 
    +     \sigma^2\s_\theta(\x, \sigma)\;,
\]
where $\s_\theta(\x, \sigma)$ is the noise conditional score function as defined in \cite{song2019generative}.
Let $w(\sigma) = \sigma^{-4}$.
Substituting these choices into the training template results in 
\begin{subequations}
\begin{align}
    \eqref{eq:template_loss} &= 
    \mathbb{E}_{\sigma, \X, \X_\sigma} \left[\sigma^{-4}\| \X - (\X_\sigma + \sigma^2\s_\theta(\X_\sigma, \sigma )) \|^2 _2\right] \\
    &=  \mathbb{E}_{\sigma, \X, \X_\sigma} \left[\sigma^{-4}\|\sigma^2\s_\theta(\X_\sigma, \sigma ) + \X_\sigma - \X \|^2_2 \right]  \\
    &=  \mathbb{E}_{\sigma, \X, \X_\sigma} \left[ \left\|\s_\theta(\X_\sigma, \sigma ) + \frac{\X_\sigma - \X}{\sigma^2} \right\|^2_2 \right] \\
    &\approx \frac{1}{L-1}  \sum_{\ell=0}^{L-1} \mathbb{E}_{\X, \X_{\sigma_\ell}} \left[
        \lambda(\sigma_\ell)\left\|
            \s_\theta(\X_{\sigma_\ell}, \sigma_\ell ) + \frac{\X_{\sigma_\ell} - \X}{\sigma_\ell^2} 
        \right\|^2_2 \right]\;,
\end{align}
\end{subequations}
where $\lambda(\sigma_\ell) > 0$ is a user-specified coefficient function depending on $\sigma_\ell$, and the derived form of training objective corresponds to the VE training loss \cite[(5-6)]{song2019generative}.

\textbf{DDPM sampling in our template.} To prove that sampling iterate of DDPM~\cite{ho_NEURIPS2020_ddpm} is a specific instance of our sampling template in~\eqref{eq:template_score},
we begin with the sampling iterates from \cite[Algorithm 2]{ho_NEURIPS2020_ddpm} and change variables to arrive at our sampling template.
We start with
\begin{equation}
\label{eq:ddpm_sampling}
    \x_{t-1} = \frac{1}{\sqrt{\alpha_t}}
    (
        \x_t - 
        \frac{1 - \alpha_t}{\sqrt{1 - \bar{\alpha}_t}}
        \bm{\epsilon}_\theta(\x_t, t)
    )
    + \sigma'_t \vec{n}\;,
\end{equation}
where $\sigma'_t$ denotes the user-defined noise schedule from \cite{ho_NEURIPS2020_ddpm},
which turns out to be distinct from the template noise schedule.

Replacing $\bm{\epsilon}_\theta(\x_t, t)$ with the score using Tweedie's formula~\cite{efron2011tweedie}:
\[
    \bm{\epsilon}_\theta(\x_t, t) = -\sqrt{1-\bar{\alpha}_t}\nabla \log f_{\X_{\bar{\alpha}_t}}(\x_t)\;,
\]
where $f_{\X_{\bar{\alpha}_t}}(\x_t)$ denotes the distribution of noisy data at time $t$, and $\X_{\bar{\alpha}_t}$ corresponds to $\X^{\text{VP}}_{t}$ in \eqref{equ:vp_noise_perturbation}. Substituting this into \eqref{eq:ddpm_sampling}, we obtain
\[
    \x_{t-1} = \frac{1}{\sqrt{\alpha_t}}
    (
        \x_t +
        (1 - \alpha_t) \nabla \log f_{\X_{\bar{\alpha}_t}}(\x_t)
    )
    + \sigma'_t \vec{n}\;.
\]
To align the state variables $\x_{t-1}$ and $\x_{t}$ on a consistent scale, we define $\s_t = \x_t/\sqrt{\bar{\alpha}_t}$. Then, we have
\[
    \sqrt{\bar{\alpha}_{t-1}} \s_{t-1} = 
     \frac{1}{\sqrt{\alpha_t}}
    (
        \sqrt{\bar{\alpha}_{t}} \s_t  +
        (1 - \alpha_t) \nabla\log f_{\vec{X}_{\bar{\alpha}_t}}(\sqrt{\bar{\alpha}_t}\s_t)
    )
    + \sigma'_t \vec{n}
\]

\[
    \s_{t-1} = 
     \frac{1}{\sqrt{\alpha_t\bar{\alpha}_{t-1}}}
    (
        \sqrt{\bar{\alpha}_{t}} \s_t  +
        (1 - \alpha_t) \nabla\log f_{\vec{X}_{\bar{\alpha}_t}}(\sqrt{\bar{\alpha}_t}\s_t)
    )
    + \frac{\sigma'_t}{\sqrt{\bar{\alpha}_{t-1}}} \vec{n}\;.
\]
Because \cite{ho_NEURIPS2020_ddpm} defines
$\bar{\alpha}_t = \prod_{s=1}^t \alpha_s$,
we have that
$\alpha_t \bar{\alpha}_{t-1} = \bar{\alpha}_t$
and therefore
\[
    \s_{t-1} =
        \s_t  +
        \frac{1 - \alpha_t}{\sqrt{\bar{\alpha}_t}} \nabla\log f_{\vec{X}_{\bar{\alpha}_t}}(\sqrt{\bar{\alpha}_t}\s_t)
    + \frac{\sqrt{\alpha_t}\sigma'_t}{\sqrt{\bar{\alpha}_{t}}} \vec{n} \;.
\]
We introduce a new index $k$ so that iterations increase, aligning with our sampling template.
We then use the score relationship in \eqref{equ:vp_score_and_ve_score} to rewrite \(\nabla\log f_{\vec{X}_{\bar{\alpha}_t}}\)
in terms of
\(\nabla\log f_{\vec{X}_{\sigma_k}}\), leading to
\[
    \s_{k+1} =
        \s_{k}  +
        \frac{1 - \alpha_{T-k}}{\bar{\alpha}_{T-k}} \nabla\log f_{\vec{X}_{\sigma_{k}}}(\s_k)
    + \frac{\sqrt{\alpha_{T-k}}\sigma'_{T-k}}{\sqrt{\bar{\alpha}_{T-k}}} \vec{n}\;.
\]
We can then rearrange it to match the form of the sampling template \eqref{eq:template_score}
\[
        \s_{k+1} =
        \s_{k}  +
        \frac{1 - \alpha_{T-k}}{\bar{\alpha}_{T-k}} \nabla\log f_{\vec{X}_{\sigma_{k}}}(\s_k)
    + \sqrt{2  \left(\frac{1 - \alpha_{T-k}}{\bar{\alpha}_{T-k}}\right)
    \left(\frac{(\sigma'_{T-k})^2\alpha_{T-k}}{2-2\alpha_{T-k}} \right)} \vec{n} \;.
\]

In \cite{ho_NEURIPS2020_ddpm},
two different choices for the sequence $\sigma'_t$ are described, with both yielding similar results.
Our analysis reveals that this sequence affects
the temperature of the sampling
and
that the temperature needs to approach one in order to correctly sample the target distribution.
In the notation of \cite{ho_NEURIPS2020_ddpm},
this requirement is
$\frac{{\sigma'}_{T-k}^{2}\alpha_{T-k}}{2-2\alpha_{T-k}} \to 1$.

Note we omit the proof for the VP-SDE sampling since the predictor-based VP-SDE iterates from~\cite[Algorithm 3]{song2021sde} are equivalent to the DDPM iterates by the proof in~\cite[Appendix E]{song2021sde}.

\textbf{NCSN sampling in our template.} Since NCSN corresponds to the same formulation as our template,
replacing time step $t$ with iteration $k$ is all we need to match between our sampling template and sampling iterates from \cite[Algorithm 1]{song2019generative}:
\[
    \x_{k+1} = \x_{k} + \frac{\varepsilon\sigma_{k}^2}{2\sigma_{\mathrm{max}}^2}\nabla \log f_{\vec{X}_{\sigma_{k}}}(\x_k) + \sqrt{\frac{\varepsilon\sigma_{k}^2}{\sigma_{\mathrm{max}}^2}}\vec{n}\;,
\]
where $\sigma_{k}$ is the designed noise level for the current state $\x_k$ and lies between $\sigma_{\mathrm{min}}$ and $\sigma_{\mathrm{max}}$, and NCSN sets $\varepsilon$ between $10^{-5}$ and $10^{-4}$\cite{song2019generative}.
Note that our sampling template follows the standard NCSN sampling iterates while allowing flexibility in step size and control over randomness through the temperature parameter.

\textbf{VE-SDE sampling in our template.} 
VE-SDE sampling \cite[Algorithm 2]{song2021sde} is already in the form of our template:
\[
    \x_{k+1} = \x_k + (\sigma_k^2 - \sigma_{k+1}^2)\nabla \log f_{\vec{X}_{\sigma_k}}(\x_k) + \sqrt{\sigma_k^2 - \sigma_{k+1}^2}\vec{n}\;.
\]
This demonstrates that our sequence of random walks framework further generalizes the SGM-SDE.

\section{Theory of Deep Score-Based Diffusion for Conditional Sampling}
\label{sec:theory_diffusion_conditional}

Building on our theoretical justification for deep score-based generative modeling,
we extend this theory to conditional sampling.
As we will show, this can be done in a straightforward way without requiring likelihood approximations.

\subsection{Existing Approaches to Conditional Sampling}
There is growing interest in adapting diffusion models for conditional sampling,
after which they can be used to solve inverse problems.
Approaches to conditional sampling with diffusion models generally follow one of two strategies:
(a) training the diffusion model conditionally on degraded measurements,
and (b) training an unconditional diffusion model and performing conditional sampling,
incorporating an additional data consistency step during the sampling process.
Strategy (a) can directly learn the posterior distribution of the desired signal given the degraded measurement, 
but requires training task-specific diffusion models, which is a data hungry and computationally expensive process~\cite{whang2022deblurconditionaldiffuion1, saharia2022srconditionaldiffusion2, liu2023dolce}.
On the other hand, strategy (b) solves the inverse problem without requiring task-specific training,
but many existing algorithms rely on ad hoc or approximate sampling strategies.
One approach implements coarse-to-fine gradient ascent on the noisy distribution by approximating the score from the residual between the output of a blind denoiser and the noisy input image~\cite{kadkhodaie2021implicit}.
Another approach~\cite{song2021sde, chung2022come, chung2022scoremri, song2022solving} alternates between sampling steps and data consistency projections. 
More recent approaches~\cite{jalal2021liklihoodapprox, kawar2021stochastic, kawar2022ddrm, chung2023dps, song2023pseudoinverse} focus on approximating the intractable time-dependent log-likelihood to achieve posterior sampling.
These approximations can be avoided if sampling is formulated in terms of a single biased random walk (as opposed to defining a forward and reverse process)~\cite{mccann2023score};
we expand upon these ideas in this work.

\subsection{Inverse Problems}
\label{subsec:inverse_problems}

Solving inverse problems involves estimating an unknown image $\x \in \mathbb{R}^d$ from its noisy measurement $\y \in \mathbb{R}^{d'}$.
The Bayesian approach to this problem is to formulate the solution
as a summary statistic (e.g., the mean or mode) of the posterior distribution, $f_{\X\mid\Y}(\x\mid\y)$.
The posterior can be expressed as the combination of a prior distribution $f_{\X}(\x)$ of desired images
and a likelihood term $f_{\Y\mid\X}(\y\mid\x)$ that describes how the known imaging system relates the unknown signal $\x$ to the  measurements $\y$
as
\begin{equation}
f_{\X\mid\Y}(\x\mid\y) \propto f_{\Y\mid\X}(\y\mid\x) f_{\X}(\x)\;.
\end{equation}
Therefore, generative models that can learn to sample directly from the posterior
or that can combine a learned prior with an analytical likelihood term during sampling
are valuable tools for solving inverse problems.

Much of the existing work on diffusion models for inverse problems focuses on linear forward models.
In this case, the
measurements $\y$ are modeled as
\begin{equation} \label{eq:inverse_problem}
    \y = \vec{A}\x + \vec{e}\;,
\end{equation}
where $\vec{A}$ is the forward operator, $\x$ is the noise-free image
and $\vec{e} \sim \mathcal{N}(\vec{0}, \eta\vec{I})$.
Since $\y$ is distributed according to $\normpdf(\vec{A}\x, \eta\vec{I})$,
the gradient of the log-likelihood is
\begin{equation} \label{equ:log_likelihood_gradient}
    \nabla\log f_{\Y\mid\X}(\y\mid\x) = \frac{1}{\eta^2}\vec{A}^{T}(\y - \vec{A}\x) \;.
\end{equation}

Unfortunately, some diffusion models require $\nabla\log f_{\Y \mid \X_{\sigma_k}}$ for conditional sampling.
This likelihood is distinct from $\nabla\log f_{\Y\mid\X}$ and typically more challenging to compute.
To estimate this, some existing works sample from posterior distribution by
training a conditional model on degraded measurements, which allows direct approximation of $\nabla \log f_{\X_{\sigma_k}\mid\Y}(\x_{k}\mid\y)$~\cite{whang2022deblurconditionaldiffuion1, saharia2022srconditionaldiffusion2, liu2023dolce}.
However, as it requires problem-specific training, leveraging pretrained unconditional diffusion model with data consistency steps is an active research area.
To approximate the intractable time-dependent log-likelihood, \cite{jalal2021liklihoodapprox} proposes
\[
\nabla\log f_{\Y\mid\X_{\sigma_k}}(\y\mid\x_{k}) \approx \vec{A}^{T}\frac{\vec{A}\x_{k}-\y}{\eta^2 + \gamma_{k}^2}\;,
\]
where $\gamma_{k}^2$ is an annealing hyper-parameter.
The class of denoising diffusion restoration models (DDRM) methods~\cite{kawar2021stochastic, kawar2022ddrm} adopts the approximation
\[
\nabla\log f_{\Y\mid\X_{\sigma_k}}(\y\mid\x_{k}) \approx \frac{\y-\x_{k}}{|\eta^2 - (\sigma_k)^2|}\;,
\]
where $\sigma_k$ denotes the noise level at diffusion iteration $k$, in the scenario where the degradation operator $\vec{A} = \vec{I}$ in \eqref{eq:inverse_problem}.
For a general $\vec{A}$, singular value decomposition is applied to weight the spectral components based on the noise level of each component.
Diffusion posterior sampling (DPS)~\cite{chung2023dps} proposes
\[
\nabla\log f_{\Y\mid\X_{\sigma_k}}(\y\mid\x_{k}) \approx \nabla\log f_{\Y\mid\X_{\sigma_k}}(\y\mid \mathsf{D}_{\theta}(\x_{k}, \sigma_k))\;,
\]
where $\mathsf{D}_{\theta}(\x_{k}, \sigma_k)$ is the MMSE reconstruction of noise-free image $\x_{0}$ given an input of $\x_{k}$ and noise level $\sigma_k$ (see \cite{chung2024reviewofdiffusion} for more comprehensive survey).
The important point from the observation of existing algorithms is that those algorithms require the approximation of the log-likelihood gradient.
In the next section,
we extend our sequence of random walks framework to solve inverse problems without any likelihood approximation.

\subsection{Sequence of Random Walks to Solve Inverse Problems}
\label{subsec:sequenceLD_solving_inverse_problems}

In Section~\ref{sec:proposed_templates},
we introduced the sequence of random walks framework as sampling template of existing diffusion algorithms.
Our theory of the sequence of random walks can easily be modified to enable sampling from a posterior distribution, thereby providing a way to solve inverse problems with a learned prior.

In the proposed framework, posterior sampling is achieved by replacing $\nabla \log f_{\X_{\sigma_k}} (\x)$ in \eqref{eq:template_score}
with $\nabla \log f_{\X_{\sigma_k}}(\x) + \nabla \log f_{\Y \mid \X}(\x)$.
The first term is expressed via a trained denoiser, while the second term is problem-specific and known in closed form.
As $\sigma_k$ approaches zero,
this sum approaches $\nabla \log f_{\X}(\x) + \nabla \log f_{\Y \mid \X}(\x)$,
which by Bayes rule is equal to the score of the desired posterior, $\nabla \log f_{\X \mid \Y}(\x)$.
This substitution results in the conditional sampling template,
\begin{align}
    \x_{k+1} &= \x_{k} + 
    \tau_k\left(
    \nabla \log f_{\X_{\sigma_k}}(\x_k)
    + \nabla \log f_{\Y \mid \X}(\x_k)\right)
    + \sqrt{2\tau_k\mathcal{T}_{k}} \vec{n} \\
    &= \x_{k} + 
    \frac{\tau_k}{\sigma^2_k} (\mathsf{D}_{\theta}(\x_k, \sigma_k) - \x_k)
    + \tau_k\nabla \log f_{\Y \mid \X}(\x_k)
    + \sqrt{2\tau_k\mathcal{T}_{k}} \vec{n}\;.
\end{align}
In the case of the linear measurement model as described above, we have
\begin{equation}
 \x_{k+1} = \x_{k} + 
    \frac{\tau_k}{\sigma^2_k} (\mathsf{D}_{\theta}(\x_k, \sigma_k) - \x_k)
    + 
    \frac{\tau_k}{\eta^{2}}
        \vec{A}^T
        \left(
        \y - 
        \vec{A}\x_{k}
        \right)
    + \sqrt{2\tau_k\mathcal{T}_{k}} \vec{n}\;.
\end{equation}
Note that since the step size $\tau_k$ is applied to both the gradient of the prior term and the gradient of the log-likelihood term,
it can be thresholded as $\min(\tau_k, L)$ to ensure known convergence requirement, where $L$ is the Lipschitz constant of log-likelihood gradient.

\begin{figure*}[t]
\begin{center}
\includegraphics[width=\textwidth]{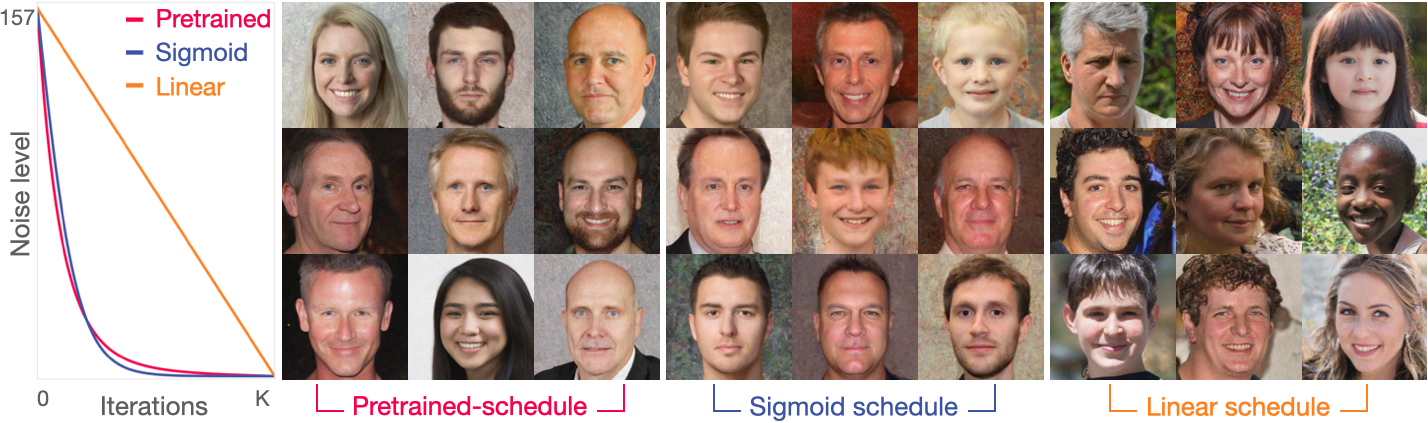}
\end{center}
\caption{Unconditional image generation with VP-score-based sequence of random walks under different noise scheduling schemes. In contrast to existing score-based diffusion sampling algorithms, which require the sampling iterations to strictly follow the noise scheduling scheme of the pretrained score function due to dependence on the Markov chain, this figure demonstrates that our framework can fully decouple from the training process by allowing $\bar{\alpha}_t$ to be defined from arbitrary noise levels $\sigma_t$, such as those following sigmoid or linear schedules.}
\label{fig:simplified_coefficient}
\end{figure*}

\section{Demonstrations of Proposed Templates}
\label{sec:numerical_validation}

While the main contribution of this work is the simple theoretical justification and unified template for diffusion model algorithms presented in Sections~\ref{sec:theory_diffusion} and \ref{sec:proposed_templates},
we now demonstrate concrete implementations of the theory.
In particular,
we show unconditional generation using simple noise schedules and straightforward parameter specifications
in a sequence of random walks framework.
We also demonstrate the effect of different temperature parameters on sample diversity and quality, highlighting how varying the ending temperature parameter influences the sampling results.
Furthermore, we empirically show this generalized and simplified approach can also be applicable to conditional generation
with the added benefit of eliminating the need for any approximation of the gradient of log-likelihood.

We now describe our experimental setup, including dataset and pretrained networks, and experiment results of our sequence of random walks framework.

\subsection{Dataset and Pretrained Neural Networks} \label{sec:data_and_networks}

For numerical validation,
we use two distinct types of pretrained diffusion models, both trained on the FFHQ 256$\times$256 face dataset \cite{karras2019ffhq}.
For the VP diffusion model, we use the one from DPS~\cite{chung2023dps}, which follows the VP training template in Table~\ref{table:template_training} with a linear schedule for $\alpha$ ranging from $\alpha_{0} = 0.9999$ to $\alpha_{T} = 0.98$.
We also adapt a pretrained VE diffusion model from SGM-SDE \cite{song2021sde},
which follows the VE training template in Table~\ref{table:template_training} with the noise level $\sigma$ varies from $\sigma_{0} = 348$ to $\sigma_{K} = 0.01$.

\subsection{Unconditional Image Sampling with Straightforward Parameter Choices}
\label{subsec:sequenceLD_straightforward}

In Section~\ref{sec:proposed_templates}, we showed that both the Markov chain theory in DDPM~\cite{ho_NEURIPS2020_ddpm}
and the reverse SDE theory in SGM-SDE~\cite{song2021sde}
can be viewed as making particular choices of step sizes $\tau_k$ and temperature parameters $\mathcal{T}_{k}$ in our framework in \eqref{eq:template_score}.
In this section,
we show that alternative, straightforward parameter choices
can provide similar unconditional sampling results.

We propose a straightforward scheme for specifying the step size and temperature term.
We define step size as a noise-level-dependent value based on the intuition that the step size should be small relative to the noise level in the current latent image $\x_k$.
Specifically,
we set $\tau_k = \varepsilon \sigma_k^2$, where $\varepsilon$ is a constant scaling factor.
Based on the theory in Section~\ref{subsec:fokker_planck},
we simplify temperature selection by setting $\mathcal{T}_0$ and linearly progressing to $\mathcal{T}_{K-1}=1$ as the final value.
This allows us to write the simplified instance within our sampling template as
\[
    \begin{aligned}
    \x_{k+1} &= \x_{k} + 
    \varepsilon\cdot(\sigma_k)^2
    \cdot
    \nabla \log f_{\vec{X}_{\sigma_k}}(\x_{k})
    + \sqrt{2\cdot \varepsilon \cdot(\sigma_k)^2\mathcal{T}_{k}} \vec{n} \\
                &= \x_{k} + 
    \varepsilon\cdot(\sigma_{k})^2
    \cdot
    \sqrt{\bar{\alpha}_{k}}
    \cdot
    \nabla \log f_{\vec{X}_{\bar{\alpha}_{k}}}(\sqrt{\bar{\alpha}_{k}} \cdot \x_{k})
    + \sqrt{2 \cdot \varepsilon \cdot(\sigma_{k})^2\mathcal{T}_{k}}  \vec{n}\;,
    \end{aligned}
\]
where \(\sigma_k = \sqrt{(1 - \bar{\alpha}_k)/\bar{\alpha}_k}\), with \(\{\sigma_k\}_{k=0}^{K-1}\) being a given sequence that implicitly determines \(\bar{\alpha}_k\).

\begin{figure*}[t]
  \begin{center}
  \includegraphics[width=0.49\textwidth]{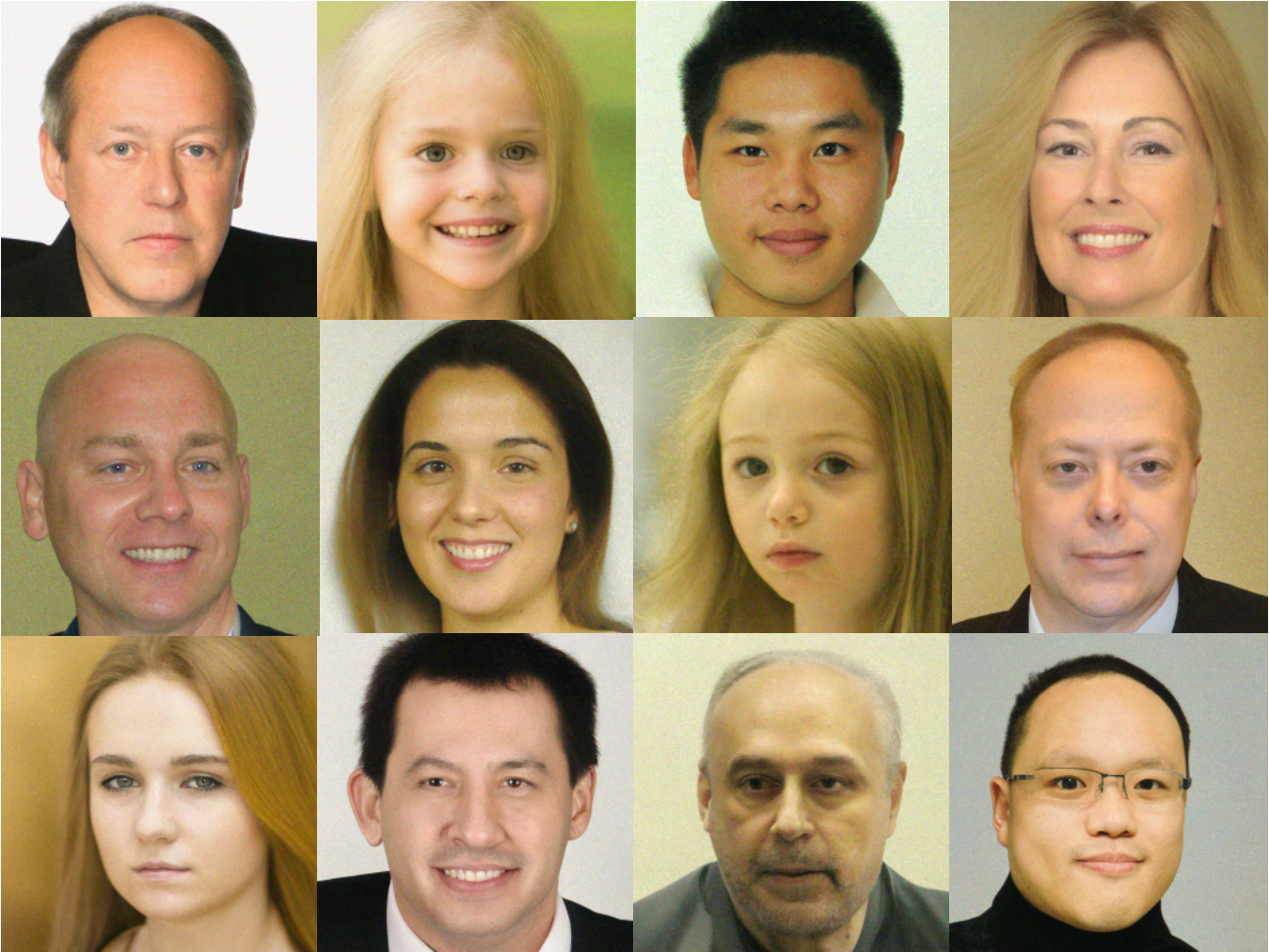}
  \end{center}
    \caption{Unconditional image generation with VE-score-based sequence of random walks.
    This figure illustrates the flexibility of our sequence of random walks framework, which enables unconditional sampling with the VE score while also being compatible with the VP score under a unified framework.
    This implies that our framework can use any type of score without restricting the score training scheme.}\label{fig:ve_sequential_langevin}
\end{figure*}

Under the proposed simplified scheme,
we compare three different choices for the noise level $\sigma_k$ in $\tau_k$ and $\nabla \log_{\X_{\sigma_k}}$.
In the first, which we call \textbf{Pretrained-schedule},
we set $\{\sigma_k\}_{k=0}^{K-1}$ to follow the same noise schedule used during the training of the pretrained score function, as outlined in Table~\ref{table:template_training}.
In the second,
which we call \textbf{sigmoid},
we set $\{\sigma_k\}_{k=0}^{K-1}$ following the specification from \cite[sigmoid schedule in Algorithm 1]{chen2023importancenoiseschedule},
matching only the initial and final noise levels to the pretrained score function's noise sequence as
\begin{equation}
    \sigma_k = \left(\sigma_0-\sigma_{K-1}\right)
    \frac{\mathcal{S}\left(\frac{c_{\text{end}}}{\zeta}\right)
    -
    \mathcal{S}\left(\frac{\frac{k}{K-1}\left(c_{\text{end}}- c_{\text{start}}\right)+c_{\text{start}}}{\zeta}\right)}
    {\mathcal{S}\left(\frac{c_{\text{end}}}{\zeta}\right)-\mathcal{S}\left(\frac{c_{\text{start}}}{\zeta}\right)}
    + \sigma_{K-1}, \quad \text{where} \quad \mathcal{S}(k) = \frac{1}{1+e^{-k}}\;.
\end{equation}
Here, $\sigma_0$ and $\sigma_{K-1}$ correspond to the noise level sequence of pretrained score function, $\zeta$ controls the stiffness of the transition, and $c_{\text{start}}$ and $c_{\text{end}}$ define the range over which the sigmoid function shapes the noise schedule.
We set variables as $(\zeta = 0.3, c_{\text{start}} = 0, c_{\text{end}} = +3)$ following \cite[first setup in Figure 3 (c)]{chen2023importancenoiseschedule}.

In the third,
which we call \textbf{linear},
we set $\{\sigma_k\}_{k=0}^{K-1}$ to linearly decrease from the maximum to the minimum of the pretrained score function's noise sequence.
Note that the second and third schemes are different from the noise schedule used during pretraining.

Given the time-conditional VP score functions rather than noise-level conditional VE score functions,
we need to find the appropriate time for the VP score to match an arbitrary noise level at each iteration.
To achieve this,
we introduce a noise-to-time mapping function $t(\sigma_k)$.
Given $\{\bar{\alpha}_t\}^{T}_{t=1}$ based on the scheduling of the pretrained VP score functions, we derive the noise sequence of VP networks as $\{\sqrt{(1-\bar{\alpha}_t)/\bar{\alpha}_t}\}_{t=1}^{T}$ as outlined in Section \ref{sec:any_denoisers}.
First, this function linearly interpolates $\{\sqrt{(1-\bar{\alpha}_t)/\bar{\alpha}_t}\}_{t=1}^{T}$ over the range $[1, 2, \ldots, K']$, where $K'$ can be much larger than number of iterations $K$ for precise matching. This function then finds $t’ = \underset{t'\in[1, K’]}{\argmin} |\sigma_{k} - \sqrt{(1-\bar{\alpha}_{t’})/\bar{\alpha}_{t’}}|$ and returns $T(\frac{t'}{K'})$ as a conditional time input of the VP score function.

For image generation using the VP diffusion model,
we take 1,000 iterations for the pretrained schedule and \textbf{100 iterations} for the sigmoid schedule
and 150,000 iterations for the linear schedule.
In order to demonstrate that visually reasonable results are achievable with each noise schedule, we tuned $\mathcal{T}_0$, $\varepsilon$, and $K$ by grid search separately for each.
For the pretrained schedule, $(\mathcal{T}_0 = 0.4, \varepsilon_{\text{pretrained}} = 0.4)$; for the sigmoid schedule, $(\mathcal{T}_0 = 0.1, \varepsilon_{\text{sigmoid}} = 0.7)$; and for the linear schedule, $(\mathcal{T}_0 = 0.95, \varepsilon_{\text{linear}} = 0.08)$.
To sample images using the VE diffusion model, we take 1,000 iterations and set
$(\mathcal{T}_0 = 0.4, \varepsilon_{\text{pretrained}} = 0.17)$.

The results of sampling using the VP diffusion model with different noise schedules (Figure~\ref{fig:simplified_coefficient})
show that plausible faces can be generated even when the noise schedule used during sampling differs from the one used during training.
The sampling results for the VE diffusion model (Figure~\ref{fig:ve_sequential_langevin}), demonstrate that the sampling template is flexible enough accommodate both VE and VP denoisers.

\begin{figure*}[t]
\begin{center}
\includegraphics[width=\textwidth]{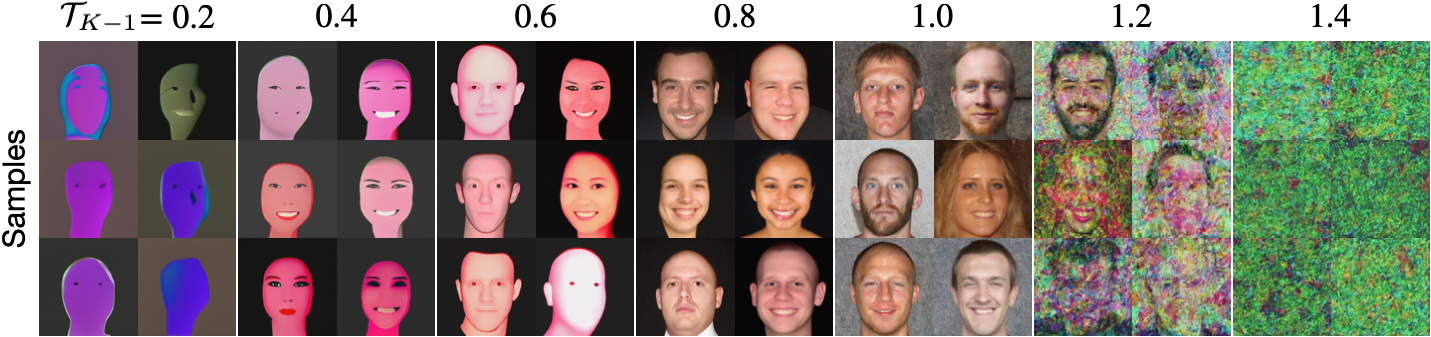}
\end{center}
\caption{Unconditional image generation using a sequence of random walks with a fixed starting temperature $\mathcal{T}_{0} = 0.4$ and varying ending temperatures $\mathcal{T}_{K-1}$.
Low $\mathcal{T}_{K-1}$ results in less variety among samples and oversmoothed faces,
high $\mathcal{T}_{K-1}$ results in noisy samples,
and $\mathcal{T}_{K-1}=1$ results in realistic samples.
}
\label{fig:different_temperature}
\end{figure*}

\subsection{Effect of Temperature Parameter} \label{subsec:temperaturecontrol}

The temperature parameter \(\mathcal{T}_k\) in our sampling template \eqref{eq:template_score} controls the stochasticity of the sampling process by scaling the noise term.
Here, we examine the effect of varying the ending temperature \(\mathcal{T}_{K-1}\) while keeping the initial temperature fixed.
To demonstrate the effect of the temperature parameter,
we follow the pretrained schedule setup described in Section~\ref{subsec:sequenceLD_straightforward}, setting the initial temperature to \(\mathcal{T}_0 = 0.4\) and linearly interpolating \(\mathcal{T}_k\) to \(\mathcal{T}_{K-1}\) over a total of 1000 iterations.
We used the step size \(\varepsilon = 0.4\) that generates plausible samples for the ending temperature \(\mathcal{T}_{K-1} = 1\). 
To analyze the impact of \(\mathcal{T}_{K-1}\), we test six different values ranging from 0.2 to 1.4. Figure~\ref{fig:different_temperature} illustrates these effects: lower temperatures yield less diverse and smoother samples, while higher temperatures introduce more noise.

\begin{figure*}[t]
\begin{center}
\includegraphics[width=\textwidth]{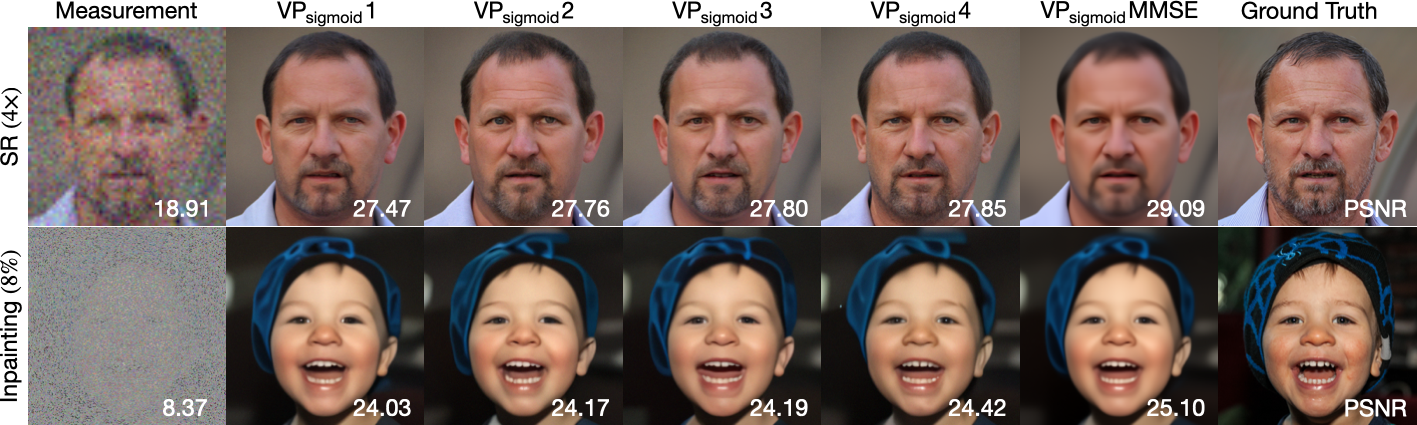}
\end{center}
\caption{Conditional image sampling results with the VP-score-based sequence of random walks with sigmoid noise scheduling.
We conditionally sample four images under the same setup, and $\text{VP}_{\text{sigmoid}}\text{MMSE}$ represents the pixel-wise average of 500 samples.
The figure illustrates that our simplified framework can extend beyond unconditional image synthesis to effectively solve inverse problems.}%
\label{fig:inverse_problems_sigmoid_vp}
\end{figure*}

\subsection{Solving Inverse Problems with Conditional Sampling} \label{subsec:experiment_inverse_euler}

We demonstrate the
conditional sampling capability of our framework on two inverse problems.
For \textbf{random inpainting}, the forward operator corresponds to randomly masking out all RGB channels of 92\% of the total pixels.
For \textbf{super-resolution},
the forward operator is bicubic downsampling by a factor of 4.
For both problems,
we set measurement noise $\eta = 0.2$.
We set step size $\tau_k = \min(\varepsilon\cdot (\sigma_{k})^2, \eta^2)$ which ensures our step size is smaller than the Lipschitz constant of log-likelihood term.
We performed sampling with both the pretrained VP and VE networks as described in Section~\ref{sec:data_and_networks}.
To sample via the VP network,
we used the sigmoid noise schedule from Section~\ref{subsec:sequenceLD_straightforward}, using $K=750$ sampling steps. The remaining parameters were specified as follows: for super-resolution $(\mathcal{T}_0 = 0, \varepsilon_{\text{sigmoid}}=0.16)$ and for inpainting $(\mathcal{T}_0 = 0, \varepsilon_{\text{sigmoid}}=0.78)$.
For the VE network, we took $K = 1,000$ sampling steps, specifying $(\mathcal{T}_0 = 0, \varepsilon_{\text{pretrained}}=0.16)$ for super-resolution and $(\mathcal{T}_0 = 0, \varepsilon_{\text{pretrained}}=0.29)$ for inpainting.

\begin{figure*}[t]
\begin{center}
\includegraphics[width=0.63\textwidth]{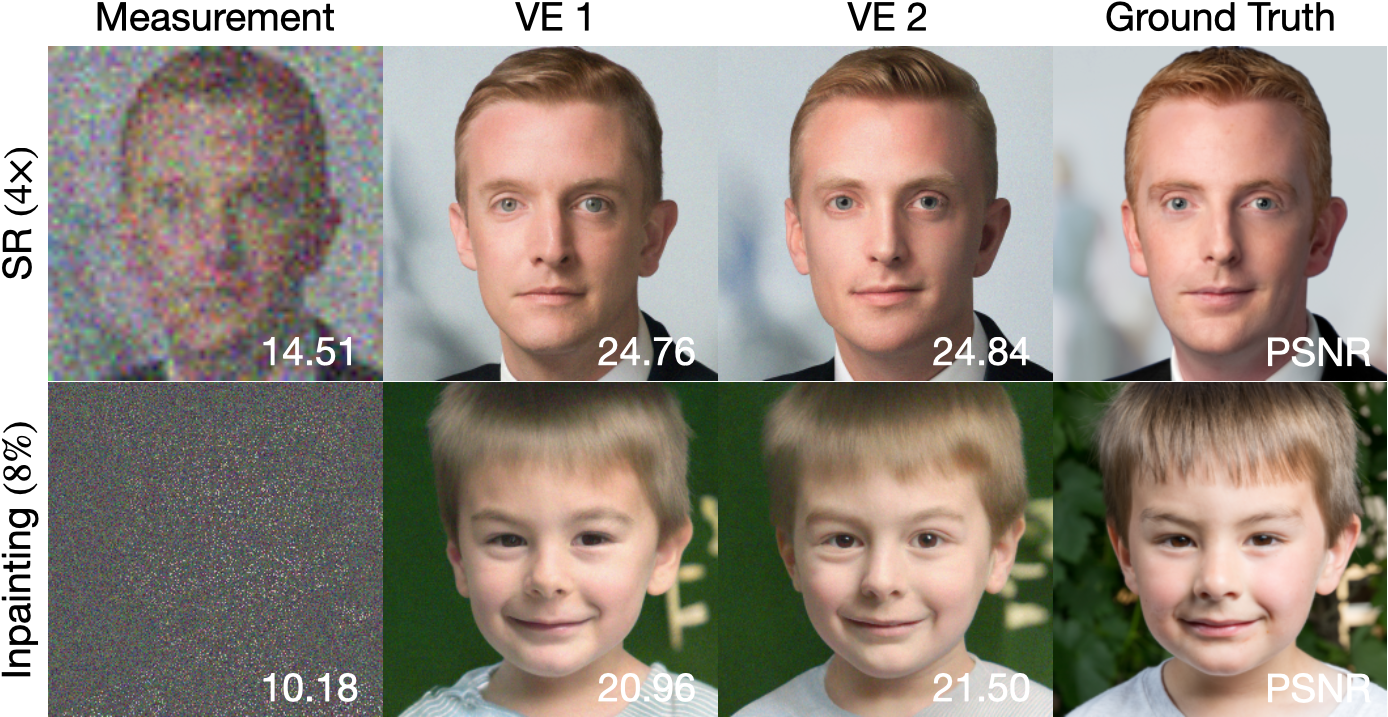}
\end{center}
\caption{Conditional sampling result with VE-score-based sequence of random walks to solve super-resolution and random inpainting.
Along with Figure~\ref{fig:inverse_problems_sigmoid_vp}, it demonstrates that the interchangeability of the VP and VE scores extends to solving inverse problems with conditional sampling.}
\label{fig:inverse_problems_ve}
\end{figure*}

Figure \ref{fig:inverse_problems_sigmoid_vp} shows results for conditional sampling with the VP network.
The four conditional samples look plausible as faces and correspond well to the measurements.
We also computed the pixel-wise average of 500 samples.
For each of these, we computed the PSNR with respect to the ground truth according to
\begin{equation}
\text{PSNR}(\x) = 10 \log_{10}\left(\frac{1}{MSE(\x_{\text{ground truth}}, \x)}\right).
\end{equation}
Training images were scaled between 0 and 1 for the VE score and between -1 and 1 for the VP score.
Sample images from the VP network were rescaled to lie between 0 and 1.
No clipping was applied to either score function, allowing all sampled values to be retained.
The PSNRs for the samples were lower than for the average result,
which makes sense because the average result should be an approximation of the MMSE reconstruction,
which maximizes PSNR on average.
Figure~\ref{fig:inverse_problems_ve} shows that the VE diffusion model also applies to our framework to solve inverse problems.

\section{Discussion}
\label{sec:Discussion}

While this work has focused on three foundational diffusion models,
many new methods have been proposed more recently.
We now comment on how some of these works may (or may not) fit into the proposed framework.
Cold diffusion replaces the stochastic diffusion process with a deterministic corruption model, where structured transformations such as blurring or downsampling degrade the data, and the learned reverse process restores the original signal without injected noise~\cite{bansal2023colddiffusion};
this approach is not readily compatible with our framework, which fundamentally relies on solving a stochastic differential equation
and on Gaussian denoising to apply Tweedie's formula.
Masked diffusion models introduce structured noise by selectively corrupting subsets of input dimensions rather than applying isotropic noise~\cite{pan2023maskeddiffusion};
this too precludes the use of Tweedie's formula.
Flow-matching approaches, which directly learn vector fields connecting a source and target distribution without requiring a stochastic diffusion process, have been explored as an alternative to score-based methods for generative modeling~\cite{lipman2023flowmatching}
and are again distinct from the theory we have presented because sampling is deterministic.
Schrödinger Bridge methods model the most likely stochastic evolution between given distributions by solving entropy-regularized optimal transport problems, leading to connections with diffusion models~\cite{debortoli2021schrodingerbridge}. 
Another recent advancement, heavy-tailed diffusion~\cite{pandey2025heavytail}, aims to improve sampling for low-probability instances by adapting alternative initial distribution such as Student's t-distribution;
our Tweedie-based framework does admit noise distributions other than Gaussian (specifically, exponential family distributions), so similar ideas could be included, but not specifically with Student's t-distribution.

\section{Conclusion}
\label{sec:Conclusion}
We have introduced a simple and self-contained theoretical justification for score-based diffusion models
that interprets sampling as solving a sequence of stochastic differential equations.
This approach provides an algorithmic template that encompasses a variety of algorithms from the literature, including NCSN~\cite{song2019generative}, DDPM~\cite{ho_NEURIPS2020_ddpm}, and SGM-SDE~\cite{song2021sde}.
The advantages of our work are that
(a) it provides a unified interpretation of influential score-based sampling algorithms;
(b) it simplifies score-based generation by decoupling the sampling noise schedule from the training noise schedule,
removing the need to define forward and reverse stochastic processes;
(c) it enables significant acceleration in sampling by allowing flexible choices of step size, temperature, and noise schedule, without being constrained by predefined heuristics;
(d) it provides algorithms for conditional sampling (i.e., solving inverse problems) without any log-likelihood approximation.
While we do not claim a new state-of-the-art for generative modeling,
we empirically demonstrate that a simple parameter and noise schedule selection within this framework can generate high-quality unconditional and conditional images.

While our framework offers a principled approach to score-based diffusion models, several theoretical and practical questions remain open.
A theoretical analysis of solvers for SDEs in the form of a sequence of random walks or annealed Langevin dynamics could refine the foundations of score-based sampling.
Additionally, studying how the smoothness of the target PDF influences sampling behavior may also provide interesting insight into the relationship between sample quality and efficiency.
The potential of deep generative modeling has barely been explored, and the signal processing community has the opportunity to make a major impact by developing new mathematical theories and practical algorithms and by finding important new applications for these fascinating, powerful techniques.

\bibliographystyle{IEEEtran}
\bibliography{refs}

\begin{thebibliography}{10}
\providecommand{\url}[1]{#1}
\csname url@samestyle\endcsname
\providecommand{\newblock}{\relax}
\providecommand{\bibinfo}[2]{#2}
\providecommand{\BIBentrySTDinterwordspacing}{\spaceskip=0pt\relax}
\providecommand{\BIBentryALTinterwordstretchfactor}{4}
\providecommand{\BIBentryALTinterwordspacing}{\spaceskip=\fontdimen2\font plus
\BIBentryALTinterwordstretchfactor\fontdimen3\font minus \fontdimen4\font\relax}
\providecommand{\BIBforeignlanguage}[2]{{%
\expandafter\ifx\csname l@#1\endcsname\relax
\typeout{** WARNING: IEEEtran.bst: No hyphenation pattern has been}%
\typeout{** loaded for the language `#1'. Using the pattern for}%
\typeout{** the default language instead.}%
\else
\language=\csname l@#1\endcsname
\fi
#2}}
\providecommand{\BIBdecl}{\relax}
\BIBdecl

\bibitem{dhariwal2021beat}
P.~Dhariwal and A.~Nichol, ``Diffusion models beat {GAN}s on image synthesis,'' in \emph{Advances in Neural Information Processing Systems}, vol.~34, 2021, pp. 8780--8794.

\bibitem{ho2022videodiffusion}
J.~Ho, T.~Salimans, A.~Gritsenko, W.~Chan, M.~Norouzi, and D.~J. Fleet, ``Video diffusion models,'' in \emph{Advances in Neural Information Processing Systems}, vol.~35, 2022, pp. 8633--8646.

\bibitem{zhang20233dsceneSOTA}
B.~Zhang, J.~Tang, M.~Niessner, and P.~Wonka, ``3dshape2vecset: A 3d shape representation for neural fields and generative diffusion models,'' \emph{ACM Transactions on Graphics (TOG)}, vol.~42, no.~4, pp. 1--16, 2023.

\bibitem{po2024sotareviewsurvey}
R.~Po, W.~Yifan, V.~Golyanik, K.~Aberman, J.~T. Barron, A.~Bermano, E.~Chan \emph{et~al.}, ``State of the art on diffusion models for visual computing,'' in \emph{Computer Graphics Forum}, vol.~43, no.~2, 2024, p. e15063.

\bibitem{chung2024reviewofdiffusion}
H.~Chung, H.~Nam, and J.~C. Ye, ``Review of diffusion models: Theory and applications,'' \emph{Journal of the Korean Society for Industrial and Applied Mathematics}, vol.~28, no.~1, pp. 1--21, 2024.

\bibitem{song2019generative}
Y.~Song and S.~Ermon, ``Generative modeling by estimating gradients of the data distribution,'' in \emph{Advances in Neural Information Processing Systems}, vol.~32, 2019.

\bibitem{vincent_connection_2011}
P.~Vincent, ``A connection between score matching and denoising autoencoders,'' \emph{Neural Computation}, vol.~23, no.~7, p. 1661–1674, Jul. 2011.

\bibitem{ho_NEURIPS2020_ddpm}
J.~Ho, A.~Jain, and P.~Abbeel, ``Denoising diffusion probabilistic models,'' in \emph{Advances in Neural Information Processing Systems}, vol.~33, 2020, pp. 6840--6851.

\bibitem{song2021sde}
Y.~Song, J.~Sohl-Dickstein, D.~P. Kingma, A.~Kumar, S.~Ermon, and B.~Poole, ``Score-based generative modeling through stochastic differential equations,'' in \emph{9th International Conference on Learning Representations (ICLR)}, 2021.

\bibitem{luo2022googlediffusiontutorial}
\BIBentryALTinterwordspacing
C.~Luo, ``Understanding diffusion models: A unified perspective,'' \emph{arXiv preprint arXiv:2208.11970}, 2022. [Online]. Available: \url{https://arxiv.org/abs/2208.11970}
\BIBentrySTDinterwordspacing

\bibitem{chan2024tutorial}
\BIBentryALTinterwordspacing
S.~H. Chan, ``Tutorial on diffusion models for imaging and vision,'' \emph{Foundations and Trends® in Computer Graphics and Vision}, vol.~16, no.~4, pp. 322--471, 2024. [Online]. Available: \url{https://doi.org/10.1561/0600000112}
\BIBentrySTDinterwordspacing

\bibitem{kingma2023understandingdiffusionELBO}
\BIBentryALTinterwordspacing
D.~P. Kingma and R.~Gao, ``Understanding diffusion objectives as the elbo with simple data augmentation,'' in \emph{Advances in Neural Information Processing Systems}, vol.~36, 2023, pp. 65\,484--65\,516. [Online]. Available: \url{https://arxiv.org/abs/2303.00848}
\BIBentrySTDinterwordspacing

\bibitem{turner2024ddpmsixsteps}
R.~E. Turner, C.-D. Diaconu, S.~Markou, A.~Shysheya, A.~Y. Foong, and B.~Mlodozeniec, ``Denoising diffusion probabilistic models in six simple steps,'' \emph{arXiv preprint arXiv:2402.04384}, 2024.

\bibitem{dockhorn2022cldsgm}
T.~Dockhorn, A.~Vahdat, and K.~Kreis, ``Score-based generative modeling with critically-damped langevin diffusion,'' in \emph{International Conference on Learning Representations (ICLR)}, 2022.

\bibitem{santos2023ornstein_ddpm}
J.~E. Santos and Y.~T. Lin, ``Using {O}rnstein-{U}hlenbeck process to understand denoising diffusion probabilistic model and its noise schedules,'' \emph{arXiv preprint arXiv:2311.17673}, 2023.

\bibitem{karras2022edm}
T.~Karras, M.~Aittala, T.~Aila, and S.~Laine, ``Elucidating the design space of diffusion-based generative models,'' in \emph{Advances in Neural Information Processing Systems}, vol.~35, 2022, pp. 26\,565--26\,577.

\bibitem{albergo2023stochasticinterpolant}
\BIBentryALTinterwordspacing
M.~S. Albergo, N.~M. Boffi, and E.~Vanden-Eijnden, ``Stochastic interpolants: A unifying framework for flows and diffusions,'' 2023. [Online]. Available: \url{https://arxiv.org/abs/2303.08797}
\BIBentrySTDinterwordspacing

\bibitem{pandey2023completerecipediffusion}
K.~Pandey and S.~Mandt, ``A complete recipe for diffusion generative models,'' in \emph{Proceedings of the IEEE/CVF International Conference on Computer Vision}, 2023, pp. 4261--4272.

\bibitem{efron2011tweedie}
B.~Efron, ``Tweedie’s formula and selection bias,'' \emph{Journal of the American Statistical Association}, vol. 106, no. 496, pp. 1602--1614, 2011.

\bibitem{mccann2023score}
M.~T. McCann, H.~Chung, J.~C. Ye, and M.~L. Klasky, ``Score-based diffusion models for bayesian image reconstruction,'' in \emph{2023 IEEE International Conference on Image Processing (ICIP)}, 2023, pp. 111--115.

\bibitem{daras2024ambienttweedie}
G.~Daras, A.~G. Dimakis, and C.~Daskalakis, ``Consistent diffusion meets tweedie: Training exact ambient diffusion models with noisy data,'' in \emph{Proceedings of the 41st International Conference on Machine Learning (ICML)}, 2024.

\bibitem{vincent2011connection}
P.~Vincent, ``A connection between score matching and denoising autoencoders,'' \emph{Neural computation}, vol.~23, no.~7, pp. 1661--1674, 2011.

\bibitem{chan_introduction_2021}
S.~H. Chan, \emph{Introduction to Probability for Data Science}.\hskip 1em plus 0.5em minus 0.4em\relax Michigan Publishing, 2021.

\bibitem{pavliotis2014langevintextbook}
G.~A. Pavliotis, \emph{Stochastic processes and applications}, ser. Texts in Applied Mathematics.\hskip 1em plus 0.5em minus 0.4em\relax Springer, 2014, vol.~60.

\bibitem{leobacher2018eulerconvergence}
G.~Leobacher and M.~Sz{\"o}lgyenyi, ``Convergence of the euler--maruyama method for multidimensional sdes with discontinuous drift and degenerate diffusion coefficient,'' \emph{Numerische Mathematik}, vol. 138, pp. 219--239, 2018.

\bibitem{debortoli2021schrodingerbridge}
V.~De~Bortoli, J.~Thornton, J.~Heng, and A.~Doucet, ``Diffusion schrödinger bridge with applications to score-based generative modeling,'' in \emph{Advances in Neural Information Processing Systems}, vol.~34, 2021, pp. 17\,695--17\,709.

\bibitem{lee2022diffusionconvergence2}
H.~Lee, J.~Lu, and Y.~Tan, ``Convergence for score-based generative modeling with polynomial complexity,'' in \emph{Advances in Neural Information Processing Systems}, vol.~35, 2022, pp. 22\,870--22\,882.

\bibitem{debortoli2022diffusionconvergence3}
\BIBentryALTinterwordspacing
V.~D. Bortoli, ``Convergence of denoising diffusion models under the manifold hypothesis,'' \emph{Transactions on Machine Learning Research}, 2022, expert Certification. [Online]. Available: \url{https://openreview.net/forum?id=MhK5aXo3gB}
\BIBentrySTDinterwordspacing

\bibitem{whang2022deblurconditionaldiffuion1}
J.~Whang, M.~Delbracio, H.~Talebi, C.~Saharia, A.~G. Dimakis, and P.~Milanfar, ``Deblurring via stochastic refinement,'' in \emph{Proceedings of the IEEE/CVF Conference on Computer Vision and Pattern Recognition}, 2022, pp. 16\,293--16\,303.

\bibitem{saharia2022srconditionaldiffusion2}
C.~Saharia, J.~Ho, W.~Chan, T.~Salimans, D.~J. Fleet, and M.~Norouzi, ``Image super-resolution via iterative refinement,'' \emph{IEEE Transactions on Pattern Analysis and Machine Intelligence}, vol.~45, no.~4, pp. 4713--4726, 2022.

\bibitem{liu2023dolce}
J.~Liu, R.~Anirudh, J.~J. Thiagarajan, S.~He, K.~A. Mohan, U.~S. Kamilov, and H.~Kim, ``{DOLCE}: A model-based probabilistic diffusion framework for limited-angle {CT} reconstruction,'' in \emph{Proceedings of the IEEE/CVF International Conference on Computer Vision}, 2023, pp. 10\,498--10\,508.

\bibitem{kadkhodaie2021implicit}
Z.~Kadkhodaie and E.~Simoncelli, ``Stochastic solutions for linear inverse problems using the prior implicit in a denoiser,'' in \emph{Advances in Neural Information Processing Systems}, vol.~34, 2021, pp. 13\,242--13\,254.

\bibitem{chung2022come}
H.~Chung, B.~Sim, and J.~C. Ye, ``Come-closer-diffuse-faster: Accelerating conditional diffusion models for inverse problems through stochastic contraction,'' in \emph{Proceedings of the IEEE/CVF Conference on Computer Vision and Pattern Recognition}, 2022, pp. 12\,413--12\,422.

\bibitem{chung2022scoremri}
H.~Chung and J.~C. Ye, ``Score-based diffusion models for accelerated {MRI},'' \emph{Medical Image Analysis}, p. 102479, 2022.

\bibitem{song2022solving}
Y.~Song, L.~Shen, L.~Xing, and S.~Ermon, ``Solving inverse problems in medical imaging with score-based generative models,'' in \emph{International Conference on Learning Representations}, 2022.

\bibitem{jalal2021liklihoodapprox}
A.~Jalal, M.~Arvinte, G.~Daras, E.~Price, A.~G. Dimakis, and J.~Tamir, ``Robust compressed sensing {MRI} with deep generative priors,'' in \emph{Advances in Neural Information Processing Systems}, vol.~34, 2021, pp. 14\,938--14\,954.

\bibitem{kawar2021stochastic}
B.~Kawar, G.~Vaksman, and M.~Elad, ``Stochastic image denoising by sampling from the posterior distribution,'' in \emph{Proceedings of the IEEE/CVF International Conference on Computer Vision (ICCV) Workshops}, 2021, pp. 1866--1875.

\bibitem{kawar2022ddrm}
B.~Kawar, M.~Elad, S.~Ermon, and J.~Song, ``Denoising diffusion restoration models,'' in \emph{Advances in Neural Information Processing Systems}, vol.~35, 2022, pp. 23\,593--23\,606.

\bibitem{chung2023dps}
\BIBentryALTinterwordspacing
H.~Chung, J.~Kim, M.~T. Mccann, M.~L. Klasky, and J.~C. Ye, ``Diffusion posterior sampling for general noisy inverse problems,'' in \emph{The Eleventh International Conference on Learning Representations}, 2023. [Online]. Available: \url{https://openreview.net/forum?id=OnD9zGAGT0k}
\BIBentrySTDinterwordspacing

\bibitem{song2023pseudoinverse}
J.~Song, A.~Vahdat, M.~Mardani, and J.~Kautz, ``Pseudoinverse-guided diffusion models for inverse problems,'' in \emph{International Conference on Learning Representations}, 2023.

\bibitem{karras2019ffhq}
T.~Karras, S.~Laine, and T.~Aila, ``A style-based generator architecture for generative adversarial networks,'' in \emph{Proceedings of the IEEE/CVF Conference on Computer Vision and Pattern Recognition}, 2019, pp. 4401--4410.

\bibitem{chen2023importancenoiseschedule}
T.~Chen, ``On the importance of noise scheduling for diffusion models,'' \emph{arXiv preprint arXiv:2301.10972}, 2023.

\bibitem{bansal2023colddiffusion}
A.~Bansal, E.~Borgnia, H.~M. Chu, J.~Li, H.~Kazemi, F.~Huang, M.~Goldblum, J.~Geiping, and T.~Goldstein, ``Cold diffusion: Inverting arbitrary image transforms without noise,'' in \emph{Advances in Neural Information Processing Systems}, vol.~36, 2023, pp. 41\,259--41\,282.

\bibitem{pan2023maskeddiffusion}
Z.~Pan, J.~Chen, and Y.~Shi, ``Masked diffusion as self-supervised representation learner,'' \emph{arXiv preprint arXiv:2308.05695}, 2023.

\bibitem{lipman2023flowmatching}
Y.~Lipman, R.~T. Chen, H.~Ben-Hamu, M.~Nickel, and M.~Le, ``Flow matching for generative modeling,'' in \emph{11th International Conference on Learning Representations (ICLR)}, 2021.

\bibitem{pandey2025heavytail}
K.~Pandey, J.~Pathak, Y.~Xu, S.~Mandt, M.~Pritchard, A.~Vahdat, and M.~Mardani, ``Heavy-tailed diffusion models,'' in \emph{13th International Conference on Learning Representations (ICLR)}, 2025.

\end{thebibliography}

\vfill

\end{document}